\documentclass[10pt, conference, compsocconf]{IEEEtran}
\usepackage{cite}

\ifCLASSINFOpdf
    \usepackage[pdftex]{graphicx}
\else
\fi
\usepackage{url}

\usepackage{booktabs}       
\usepackage{nicefrac}       

\begin{document}
%
\title{The Ciona17 Dataset for Semantic Segmentation \\
of Invasive Species in a Marine Aquaculture Environment}

\author{
	\IEEEauthorblockN
	{
		Angus Galloway\IEEEauthorrefmark{1},
		Graham W. Taylor\IEEEauthorrefmark{1},
		Aaron Ramsay\IEEEauthorrefmark{2},
		Medhat Moussa\IEEEauthorrefmark{1}
	}
	\IEEEauthorblockA
	{
		\IEEEauthorrefmark{1}
		School of Engineering\\
		University of Guelph\\
		Guelph, ON, Canada\\
	    \{gallowaa, gwtaylor, mmoussa\}@uoguelph.ca\\
	    \\\IEEEauthorrefmark{2}
	    Department of Agriculture and Fisheries\\
	    Government of PEI\\
	    Montague, PEI, Canada\\
	    \{apramsay@gov.pe.ca\}\\
	}
}


%


\maketitle

\begin{abstract} An original dataset for semantic segmentation, ``Ciona17'', is
introduced, which to the best of the authors' knowledge, is the first dataset of
its kind with pixel-level annotations pertaining to invasive species in a marine
environment. Diverse outdoor illumination, a range of object shapes, colour, and
severe occlusion provide a significant real world challenge for the computer
vision community. An accompanying ground-truthing tool for superpixel labeling,
``Truth and Crop'', is also introduced. Finally, we provide a baseline using a
variant of Fully Convolutional Networks, and report results in terms of the
standard mean intersection over union (mIoU) metric.

\end{abstract}


%
\IEEEpeerreviewmaketitle

\section{Introduction}
In long-line mussel farming, the fouling of mussel socks and gear by
\emph{tunicates}, a marine invertebrate species, can impede mussel growth,
decrease yield, and reduce overall farm productivity
\cite{ramsay-ciona-pei,colonial-vs-ciona,ciona-treatment,netherlands}.
\emph{Ciona intestinalis} are widely
considered to be the most problematic of the tunicates in mussel farming
due to their rapid proliferation and biomass
\cite{ramsay-ciona-pei,ciona-treatment}. Aquaculture industries in
The Netherlands, New Zealand, and particularly Prince Edward Island (PEI),
Canada, are plagued by \emph{Ciona}; having the ability to displace other
invasive tunicates, such as \emph{Styela clava}
\cite{ramsay-ciona-pei,netherlands}.\par

\begin{figure}[h]
	\centering
	\includegraphics[width=0.95\linewidth]{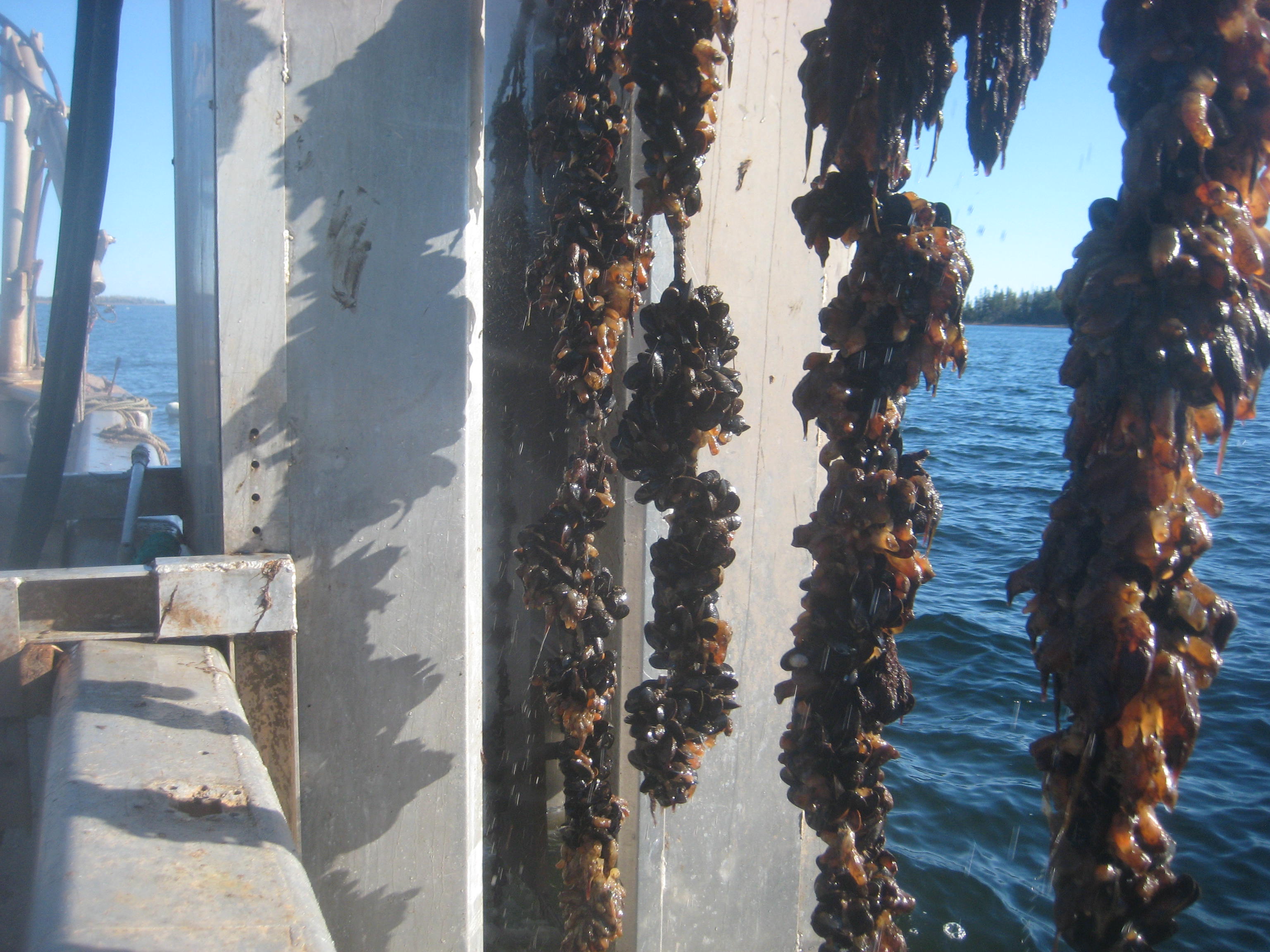}
	\caption{Long-line farmed mussel socks with moderate \emph{Ciona}
		infestation (orange gelatinous specimen) about to enter a high-pressure
		treatment system fixed to hull of vessel. A generic design for this system
		is available \cite{PEIAA}, but each farm's equipment is made to order.
		Best viewed in colour.}
	\label{fig:spray-treatment-example}
\end{figure}

In response to the \emph{Ciona} problem, an above-water high-pressure treatment
system, depicted in Figure~\ref{fig:spray-treatment-example}, has been in use
for several years \cite{ciona-treatment}, but this design has several
significant limitations.
Current mitigation systems do remove substantial \emph{Ciona} biomass, but they
also introduce avoidable mussel fall-off, as the force of water jets can exceed
the tensile strength of the mussels' byssal threads. \par

The fall-off problem is further exacerbated during the summer months, when
byssal threads are weaker \cite{byssal-thread-mechanics}, spawning of
\emph{Ciona} is at peak intensity \cite{ramsay-ciona-pei}, and treatments are
most frequent. Furthermore, mussel fall-off is expected to increase
with rising ocean acidification \cite{ocean-acidification}.\par

The high-pressure treatment system can be considered open-loop with respect to
the extent of infestation on a mussel sock. Quantifying invasive
species by surface area, and location, in real-time, is central to the
feasibility of an improved closed-loop computer vision-based treatment system.
Such a system, paired with electromechanical nozzles, could
strategically target invasive species, minimizing mussel fall-off as well as
the number of requisite treatments in a growing season.

Measuring the amount of biofouling on mussel socks before and after
treatment, is also desirable for making informed management decisions. Given a
cleanliness metric, farmers can counteract inconsistent treatments caused by
the use of non-standardized equipment (e.g.~different nozzles, orientation,
water pressure), and varying treatment speeds. \par

The farming context is the motivation for formulating the problem as one of
\emph{semantic segmentation} as opposed to simply \emph{object detection},
since we seek to describe \emph{where}, \emph{what kind}, and \emph{how much}
infestation is present. Counting individual object instances is of less
importance in mitigation, but may be of scientific interest in other
capacities, such as a detailed study of recruitment patterns
\cite{colonial-vs-ciona}.\par

\section{Related Work}

In the context of semantic segmentation, to date, several major datasets have
served as common benchmarks for assessing algorithm performance. Yet,
comparatively few arise from real-world, application specific problems. In this
section we discuss some alternative datasets that facilitate supervised semantic
segmentation with pixel-level annotations, and draw comparisons with Ciona17.

\subsection{PASCAL Visual Object Classes
\cite{pascal-voc-2012}}

The most recent PASCAL Visual Object Classes (VOC) challenge dataset has 21
object classes (e.g.~aeroplane, bicycle, bird), and provides 2,913 pixel-level
segmentations \cite{pascal-voc-2012}.

The PASCAL dataset addressed several criticisms of earlier object recognition
datasets, which lacked clutter and diverse object orientations. Given that the
PASCAL images were drawn pseudo-randomly from the consumer photo sharing website
Flickr, they tend to be representative of challenging real-world scenes.
Although the PASCAL VOC competitions have ended, the datasets continue to
be a standard benchmark, having led to
several breakthroughs in the field of deep learning for semantic segmentation
\cite{DBLP:journals/corr/LongSD14,DBLP:journals/corr/NohHH15,CP2015Semantic}.

\subsection{The Cityscapes Dataset for Semantic Urban Scene Understanding
\cite{Cordts2016Cityscapes}}

This more modern dataset extends PASCAL VOC to high-complexity urban scenes,
and features 5,000 fine pixel-level annotations, and 20,000 self described
``coarse'' annotations, for 30 different object classes (e.g.~person,
rider, car, truck) and eight broader categories (e.g.~human, vehicle, nature).
For testing, only 18 of the sufficiently unique objects are used to assess
performance.
The dataset organizers employed two forms of quality control regarding the
annotations: 1) having 30 images labeled by two different annotators, and 2)
coarse labeling of previously fine labeled images. Both measures resulted in
greater than 96\% overlap, a testament to high consistency even among numerous
annotators.

\subsection{Comparisons}

Similarly to the Ciona17 dataset, each example in PASCAL consists of an RGB
image, and segmentation mask containing integer indices corresponding to an
object class. Unlike the PASCAL dataset, Ciona17 does not contain object
instance masks, as the practical nature of farming does not require counting of
individual species. Furthermore, a single cropped image of modest resolution
could potentially contain hundreds of individual \emph{Ciona} or mussels,
making this impractical to label accurately.

In comparing the PASCAL dataset and Ciona17, the latter contains even more
occlusion and blending of classes, as the tunicates tend to grow directly on top
of, or among, the mussels. The PASCAL dataset does contain some occlusion, such
as ``person riding a bicycle'' or ``busy street scene'', however many examples
feature one or two instances of an unobstructed object, e.g.~``aeroplane in the
sky''. In comparison, many randomly oriented object instances with heavy
occlusion, is the norm rather than the exception, in Ciona17.

The Cityscapes dataset is more similar to Ciona17, as they are both
application-specific and most examples contain several crowded objects, with
some Cityscapes scenes having over a hundred instances. Cityscapes is also
highly occluded, with vehicles and pedestrians obstructing views of other
people.

\section{The Ciona17 Dataset}

We introduce a ground-truthed dataset derived from high-resolution images taken
on two different mussel farms\footnote{http://dx.doi.org/10.5683/SP/NTUOK9}.
Images were collected with a consumer-grade Canon PowerShot SD1000 7.1MP
digital camera mounted on a tripod with ISO fixed to 100. Both farms were
located in eastern PEI and experienced significant infestations
of \emph{Ciona intestinalis}. Only one of the farms, ``Farm 2'', had visible
\emph{Styela clava}. \par

No specific instruction was given to the farmers regarding
the type of crop to show. They were simply told to treat what they had
originally intended to do that day. We have high confidence that the data
collected is representative of what a typical farm would encounter during a
normal treatment cycle. Data collection occurred on two consecutive days near
the end of the tunicate spawning season in PEI, in early November 2016.
This was the last scheduled day of treatments for Farm 1, as \emph{Ciona} cease
to reproduce below water temperatures of about $8^{\circ}C$
\cite{ramsay-ciona-pei}.\par

\subsection{High Level Overview}

The Ciona17 dataset consists of 1,472 images with corresponding pixel-level
segmentations. The images are cropped to a uniform $224\times224$px, and the
following semantic labels were assigned and stored to integer masks:

\begin{itemize}
	\item \textbf{0 - Other}
	\item \textbf{1 - Mussel}
	\item \textbf{2 - Ciona}
	\item \textbf{3 - Styela}
	\item \textbf{4 - Void}
\end{itemize}


The ``Other'' class represented uninteresting background classes that were not
explicitly annotated, while ``Void'' was used for regions that were difficult
to label (e.g.~foreground class overlap).
The distribution of class labels is shown in Table~\ref{class-distribution-f1}.
The dataset is split into two training, and one test set. The test
set has significantly more \emph{Ciona}, and facilitates testing how resulting
algorithms handle class-imbalance. Both training sets have a similar
volume of \emph{Ciona}, but Training 2 was unique in receiving a void label
during annotation, in similar fashion to PASCAL VOC. We are aware of additional
complexities in adapting some models to use void labels, such as Fully
Convolutional Networks (FCN) \cite{DBLP:journals/corr/LongSD14} implemented in
frameworks like TensorFlow \cite{tensorflow2015-whitepaper}, therefore we keep
Training 2 separate.

\begin{table}[h]
	\caption{Class balance for training and test data for Farm 1.}
	\label{class-distribution-f1}
	\centering
	\begin{tabular}{lccccc}
		\toprule
		Split & Images & Other \% & Mussel \% & Ciona \% & Void \%\\
		\midrule
		Training 1 & 702 & 38 & 49 & 13 & 0\\
		Training 2 & 364 & 35 & 29 & 14 & 19\\
		Test & 334 & 42 & 18 & 40 & 0\\
		\bottomrule
	\end{tabular}
\end{table}

\subsection{Truth and Crop Tool}

To facilitate quick ground truthing of full size images, a platform-independent
GUI utility ``Truth and Crop'' was implemented in
Qt\footnote{https://github.com/AngusG/truth-and-crop}. This tool let the
annotator label whole superpixels, rather than individual pixels, or by drawing
polygons. Using the tool, an annotator performs an initial ground truth, and
can then click arbitrarily to save cropped image-segmentation mask pairs.

\begin{figure}[h]
	\centering
	\includegraphics[width=0.95\linewidth]{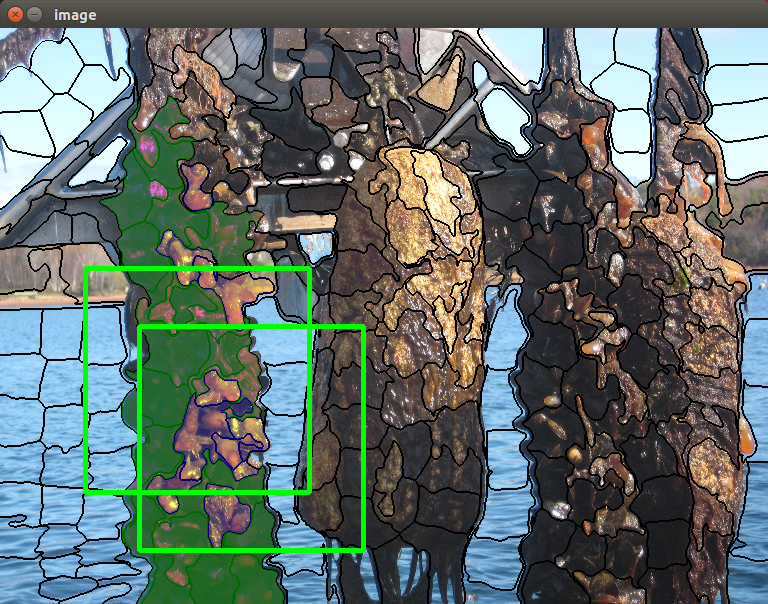}
	\caption{Screenshot of ``Truth and Crop'' utility with superpixel segments
	extracted via SLIC \cite{scikit-learn} for full size image downsampled by
	four (4), with $\sigma=3$, and 300 roughly equal sized segments. Two mussel
	socks, and two Styrofoam buoys are present, with lime-green bounding boxes
	where cropped regions are to be extracted. (\emph{Left}) Ground
	truthed regions of sock have a transparent colour fill, with green for
	1-Mussel, and purple for 2-Ciona. Regions that are not filled are
	considered 0-Other after cropping operation.
	(\emph{Right}) Image not ground truthed to show raw segments. Best viewed
	in colour.}
	\label{fig:truth-and-crop}
\end{figure}

The simple linear iterative clustering (SLIC) superpixel extraction
algorithm \cite{slic} is used to automatically find boundaries around similar
parts or edges in the image. The SLIC algorithm implemented in the Python
scikit-image \cite{scikit-learn} library receives as arguments the number of
roughly equal size segments requested, width of Gaussian smoothing kernel,
$\sigma$, and a segment compactness factor. Once superpixel clusters have been
identified, they can be imposed on the original image, and the annotator
assigns a class label to all pixels in a superpixel with a single click. This
process is depicted in Figure~\ref{fig:truth-and-crop}.

\subsection{Ground Truthing Procedure}

Images were ground truthed by the primary author after meeting with
aquaculture biologists, and spending significant time manually inspecting live
mussel socks.

Through the Truth and Crop GUI, SLIC hyperparameters were set manually on a
per-image basis, while spending roughly 10-15 minutes per image for the whole
annotation procedure. An optional down-sampling factor, usually of no more than
two or three,  was first applied to the original image in both height and width
dimensions. This helped ensure that samples were drawn from a variety of
scales, and made the labeling process faster. After labeling foreground
classes, all other pixels were automatically assigned to the ``Other'' class.

\begin{figure}[h]
	\centering
	\includegraphics[width=0.95\linewidth]{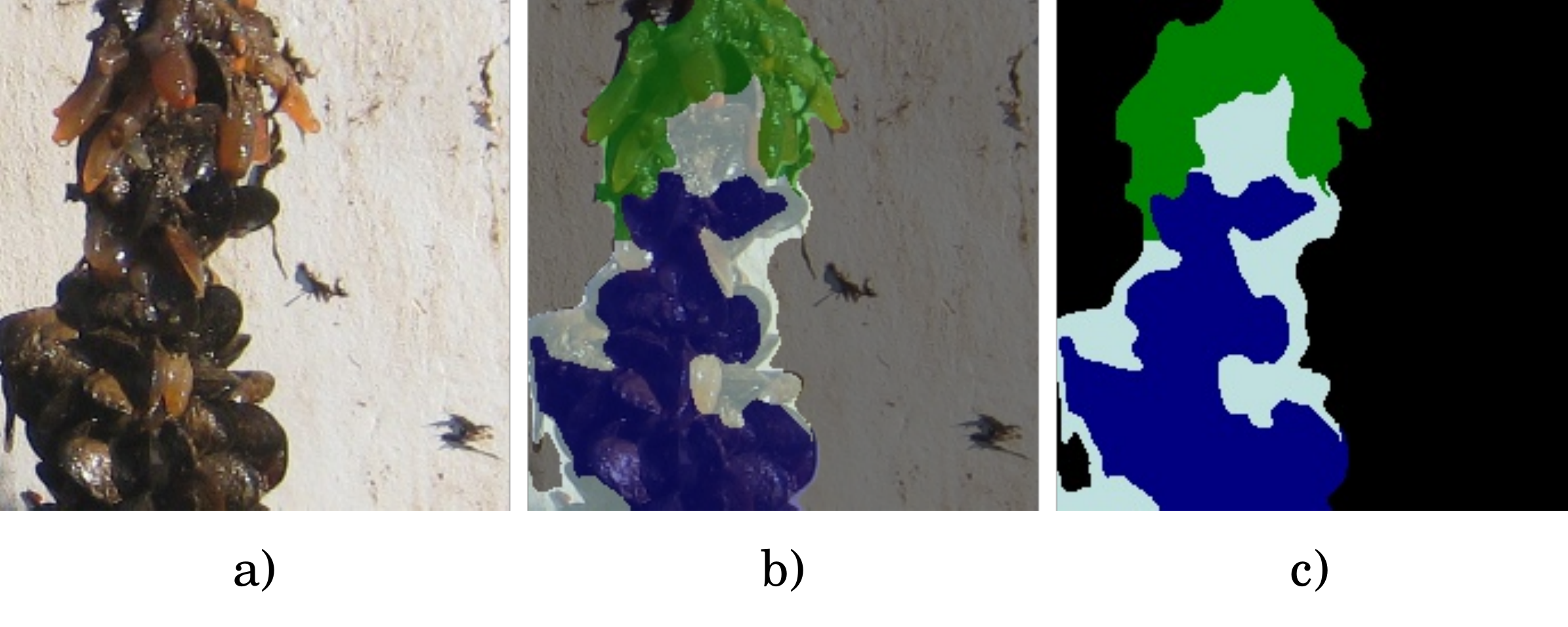}
	\caption{Ciona17 sample of mussel sock top section from Farm 1. \emph{a)}
		RGB image, \emph{b)} segmentation mask imposed on \emph{a)}, and
		\emph{c)} segmentation mask with \emph{Ciona} as green area, mussels as
		blue area, and void as white area. Best viewed in colour.}
	\label{fig:dataset_sample}
\end{figure}

Figure~\ref{fig:dataset_sample} depicts some of the challenging decisions that
had to be made during the annotation process. Some difficult images
resulted in superpixels that contained several classes, e.g.~background and a
mussel, or mussel and \emph{Ciona}. In general, for cases where the annotator
estimated that each class was roughly equally represented in a superpixel, the
void label was assigned. Since all labels had to be assigned to a whole
superpixel, the void label is more prominent than the thin border line that
appears in PASCAL, however this was still considered efficient compared to
manually drawing polygons with alternative tools such as LabelMe
\cite{Russell08labelme:a}. \par

The majority of superpixels resulted in much sharper edges in well illuminated
areas than would have been feasible to draw in LabelMe. Superpixels that
contained only \emph{Ciona} and mussel were assigned the majority class. The
void label was also used in some cases that were too difficult to discern,
usually in areas with particularly low light.

To ensure good coverage of annotated regions, there was some overlap between
images, similarly to what can be observed in Figure~\ref{fig:truth-and-crop}.
We emphasize that for the splits outlined in Table~\ref{class-distribution-f1},
there was no overlap between the training and test sets as these images
were cropped from different original images.

\subsection{Dataset Features}

We explained previously how the Ciona17 dataset relates to open
problems in aquaculture. Now, a case is made for Ciona17 as a challenging
and interesting computer vision dataset in general.

\subsubsection{Variable Lighting}

\begin{figure}[h]
	\centering
	\includegraphics[width=0.95\linewidth]{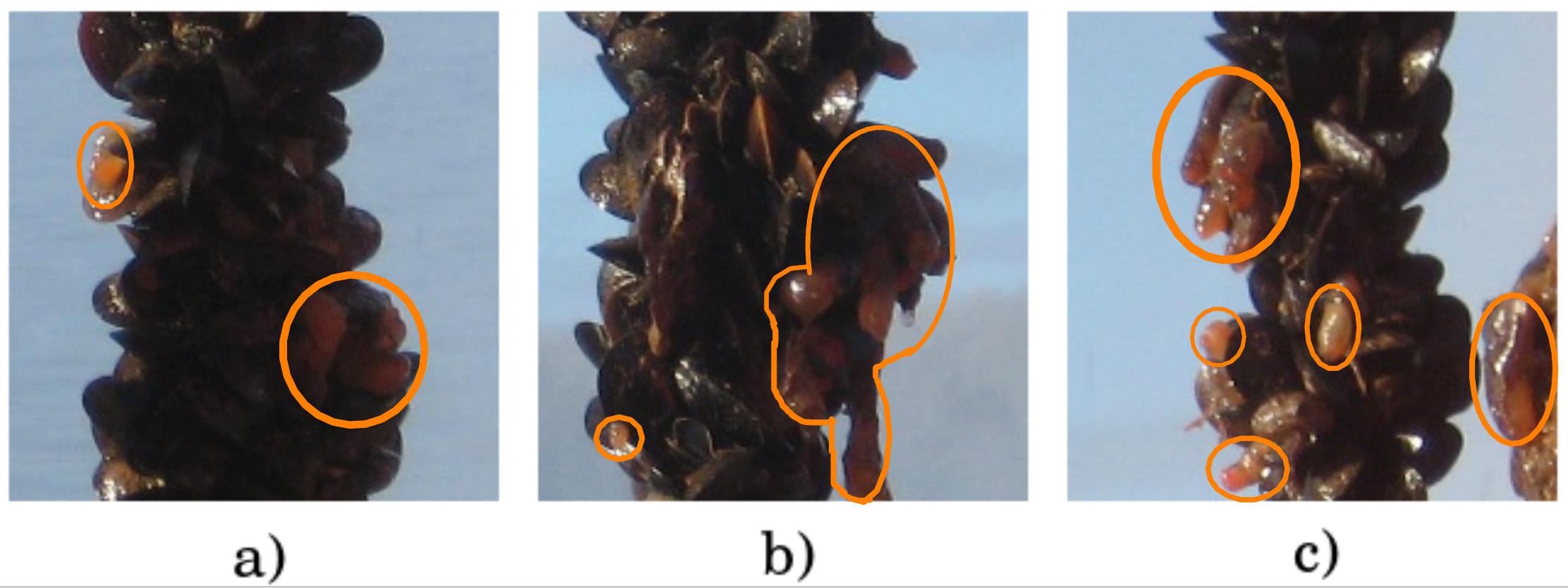}
	\caption{Samples of \emph{a)} ``low'', \emph{b)} ``medium'', and \emph{c)}
	``high'' illumination sock sections taken from the same original image.
	Sections \emph{a)} and \emph{b)} are from the same sock, while \emph{c)} is
	from an adjacent sock. All instances of \emph{Ciona} were carefully
	annotated in orange.
	Best viewed in colour.}
	\label{fig:low-med-high-hist}
\end{figure}

Although the weather was clear and sunny during both days that images were
collected, sock sections cast in shadow make accurate segmentation
difficult, even for the human eye. Figure~\ref{fig:low-med-high-hist} shows
how challenging it can be to identify clusters of \emph{Ciona} on mussel socks
in broad daylight, due to a range of illumination and shadow inherent in
outdoor imagery.

\subsubsection{Colour}

\begin{figure}[h]
	\centering
	\includegraphics[width=0.95\linewidth]{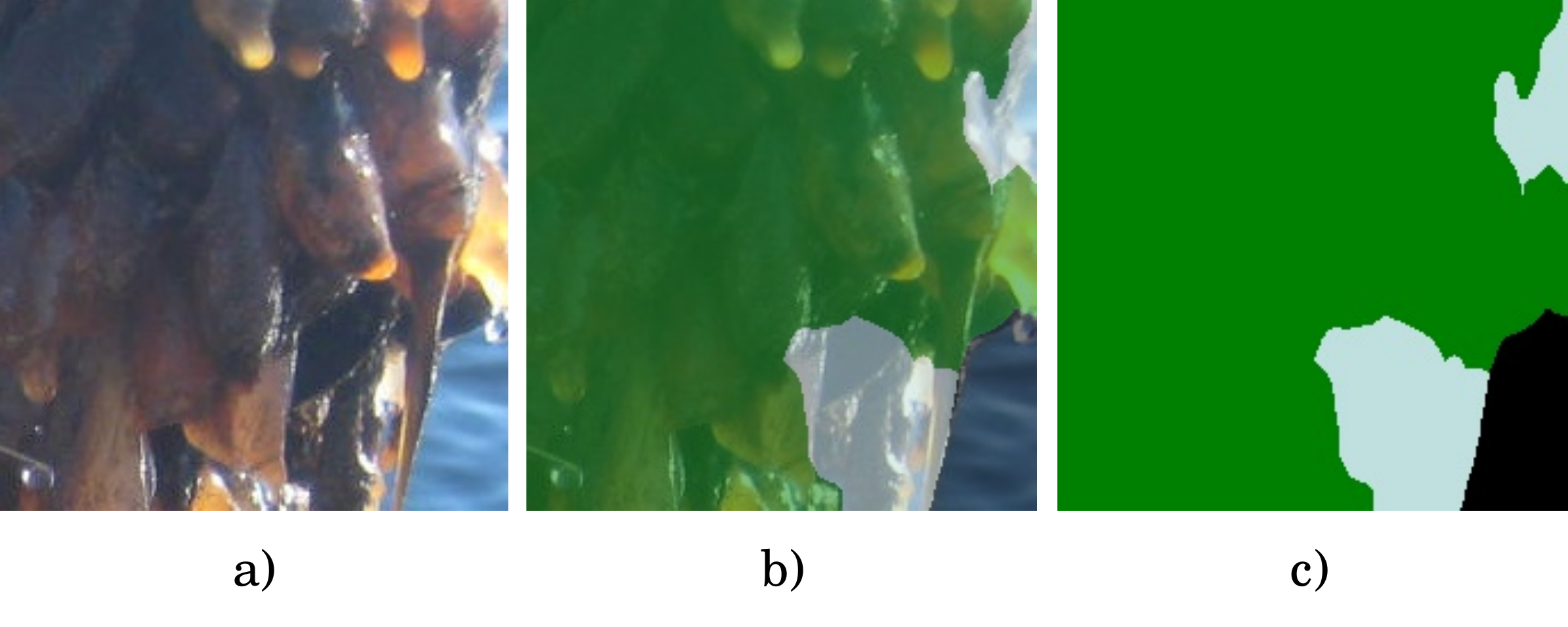}
	\caption{Ciona17 sample of predominantly \emph{Ciona} with diverse colour
		from Farm 2. \emph{a)} RGB image, \emph{b)} segmentation mask imposed
		on \emph{a)}, and \emph{c)} segmentation mask with \emph{Ciona} as
		green area, and void as white area. Two superpixels had to be voided
		due to mussel grouped together with debris and background. Best viewed
		in colour.}
	\label{fig:ciona_diverse_colour}
\end{figure}

In theory, colour thresholding should separate mussels from the mostly
bright orange \emph{Ciona}, but this assumes direct sunlight, and lack of
silt or debris. It has also been speculated that \emph{Ciona} tend to
darken with age \cite{aaron-email-ciona-colour-age}. The sample in
Figure~\ref{fig:ciona_diverse_colour} shows a typical spectrum of colour for
\emph{Ciona}.

Additionally, some mussel shells reflect a similar orange colour to that of
\emph{Ciona}, an effect visible in Figure~\ref{fig:low-med-high-hist} \emph{b)}
and \emph{c)}. This tends to occur naturally around the seam where the shell
halves meet, but silt also contributes to a more global effect. \par

\subsubsection{Texture}

\emph{Styela clava} are the mostly highly textured of all species in the
dataset, but their colour is less distinguishable from mussels. Older
\emph{Ciona} have a finely pitted texture as their tunic thickens, while
\emph{Styela} consistently possess a coarse sandpaper-like appearance.

\begin{figure}[h]
	\centering
	\includegraphics[width=0.55\linewidth]{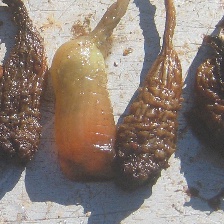}
	\caption{Sample of individual \emph{Ciona} and \emph{Styela} from Farm 2.
	\emph{Left} \emph{Ciona} and \emph{Right} \emph{Styela}. Best viewed in
	colour.}
	\label{fig:ciona_styela_texture}
\end{figure}

\subsubsection{Occlusion}

Farm 2 employed a technique known as ``double socking''
in which mussel socks are wrapped in-place with an additional mesh to
prevent fall-off. This mesh can be considered as added occlusion in the
dataset and is a common practice on some farms. The double socking made precise
annotation extremely difficult, therefore we have held out this data for
future consideration (e.g.~for use in unsupervised or semi-supervised learning).

\section{Evaluating Performance}

To penalize naive schemes that optimize for background detection, we suggest
using the standard ``Mean Intersection over Union'' (mIoU) metric, commonly
used in semantic segmentation \cite{Everingham2015}. For a given class, and
true positives (TP), false positives (FP), and false negatives (FN), the
intersection over union (IoU) is given by:

\begin{equation}
IoU = \frac{TP}{TP+FP+FN}
\end{equation}

Averaging the IoU scores for each class results in the overall mIoU score. In
this way, a scheme that overwhelmingly predicts background will have little
intersection with masks for mussels or \emph{Ciona}, as no credit is earned for
true negatives.

\subsection{Baseline Result}

To establish a baseline mIoU score on the Ciona17 dataset, a variant of FCN
\cite{DBLP:journals/corr/LongSD14} outlined in 
Table~\ref{tab:VGG6S-Fc6-512-Deconv}, was trained end-to-end with
pixelwise softmax cross-entropy loss and ReLU activations. 

\begin{table}[h]
	\caption{Fully Convolutional Network architecture (VGG6S-Fc6-512-Deconv) used
  as our baseline.}
	\label{tab:VGG6S-Fc6-512-Deconv}
	\centering
	\begin{tabular}{llcl}
		\toprule
		Layer Type & Kernel & Stride & Output Shape \\
		\midrule
		Input & n/a & n/a & 224$\times$224$\times$3 \\
		\midrule
		Conv\_1 & 3$\times$3 & 1 & 224$\times$224$\times$16 \\
		Pool\_1 & 2$\times$2 & 2 & 112$\times$112$\times$16 \\
		\midrule
		Conv\_2 & 3$\times$3 & 1 & 112$\times$112$\times$32 \\
		Pool\_2 & 2$\times$2 & 2 & 56$\times$56$\times$32 \\
		\midrule
		Conv\_3 & 3$\times$3 & 1 & 56$\times$56$\times$64 \\
		Pool\_3 & 2$\times$2 & 2 & 28$\times$28$\times$64 \\
		\midrule
		Conv\_4 & 3$\times$3 & 1 & 28$\times$28$\times$128 \\
		Pool\_4 & 2$\times$2 & 2 & 14$\times$14$\times$128 \\
		\midrule
		Conv\_5 & 3$\times$3 & 1 & 14$\times$14$\times$256 \\
		Pool\_5 & 2$\times$2 & 2 & 7$\times$7$\times$256 \\
		\midrule
		Fc\_6 & 1$\times$1 & 1 & 7$\times$7$\times$512 \\
		Deconv\_1 & 64$\times$64 & 32 & $ 224 \times 224 \times 3$ \\
		\bottomrule
	\end{tabular}
\end{table}

\begin{figure}[h]
	\centering
	\includegraphics[width=0.95\linewidth]{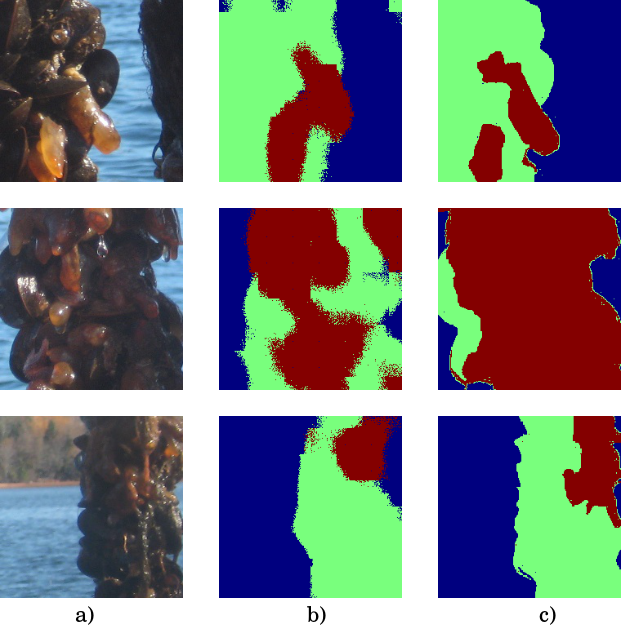}
	\caption{Three sets of \emph{a)} RGB image sample from Ciona17 test set,
	\emph{b)} prediction of trained VGG6S model, and \emph{c)} segmentation
	mask with \emph{Ciona} as red, mussels as green, and other as blue. Best
	viewed in colour.}
	\label{fig:vgg6s_benchmark_samples}
\end{figure}

Training was performed with 
the adaptive moment estimation (Adam) gradient descent scheme 
\cite{DBLP:journals/corr/KingmaB14}, initial learning rate of $1 \times 
10^{-5}$, mini-batch size of 10, and dropout with a dropout rate of 0.5 on 
layer \texttt{Fc\_6}. The RGB images were not preprocessed other than a simple 
conversion of the 8-bit pixel intensities to floating point values in the range 
[0-1]. Training was interrupted at 100k steps, re-starting the optimizer from
scratch, and allowed to continue until 300k total steps had elapsed 
resulting in $mIoU_{test} = 51.36\%$. Some samples that were drawn from the 
trained model are shown in Figure~\ref{fig:vgg6s_benchmark_samples}. 

In practice, it may be acceptable to merge the ``other'' and ``mussel''
classes, turning the problem into that of foreground/background binary
segmentation, as a closed loop treatment system only needs to distinguish
\emph{Ciona} from everything else. Despite this, farmers may wish to
contextualize the \emph{Ciona} measure with mussel biomass, hence the need to
report IoU across all classes.

\section{Conclusion}

To the best of the authors' knowledge, Ciona17 is the first dataset of its
kind with pixel-level annotations pertaining to invasive species in a marine
environment. It is the authors' intent that in making this dataset available,
the research community will propose new models that are particularly adept
at controlling for highly variable illumination and occlusions, exceeding the
initial benchmark mIoU of 51.36\%.

In making the original images and tools used for annotation available, we
expect that improved ways of annotating challenging datasets such as
this one will be proposed.

\section*{Acknowledgment}

The authors thank Brian Gillis of the \emph{PEI Department of Agriculture and
Fisheries} for assisting with the data acquisition. We also thank Colin
Reynolds of \emph{Reynold's Island Mussels}, and Dana Drummond's crew at
\emph{Atlantic Aqua Farms} for offering their farms for data collection.



%

\bibliography{master}
\bibliographystyle{ieeetr}

\end{document}